\documentclass[letterpaper]{article} % DO NOT CHANGE THIS
\usepackage[preprint]{aaai2027}  % DO NOT CHANGE THIS
\usepackage[hyphens]{url}  % DO NOT CHANGE THIS
\usepackage{graphicx} % DO NOT CHANGE THIS
\usepackage{natbib}  % DO NOT CHANGE THIS AND DO NOT ADD ANY OPTIONS TO IT
\usepackage{caption} % DO NOT CHANGE THIS AND DO NOT ADD ANY OPTIONS TO IT
\usepackage{algorithm}
\usepackage{algorithmic}

\usepackage{newfloat}
\usepackage{listings}
\DeclareCaptionStyle{ruled}{labelfont=normalfont,labelsep=colon,strut=off} % DO NOT CHANGE THIS
\floatstyle{ruled}
\newfloat{listing}{tb}{lst}{}
\floatname{listing}{Listing}

\usepackage{booktabs}

\usepackage{amssymb}
\usepackage{tikz}
\newcommand*\circled[1]{\tikz[baseline=(char.base)]{
		\node[shape=circle,draw,inner sep=0.2pt] (char) {#1};}}
\newcommand*\circledB[1]{\tikz[baseline=(char.base)]{
        \node[shape=circle,fill,inner sep=0.2pt] (char) {\textcolor{white}{#1}};}}
\newcommand{\add}[1]{\textcolor{black}{#1}}

\usepackage{comment}

\title{ENCL: Enabling Efficient Neuromorphic Continual Learning with Local Learning Rule and Sub-Network Freezing Strategy}
\author{
    Rachmad Vidya Wicaksana Putra, 
    Shuochen Zhou, 
    Hassan Raza, 
    Muhammad Shafique
}
\affiliations{
    eBRAIN Lab, New York University (NYU) Abu Dhabi, United Arab Emirates\\
    \{rachmad.putra, sz4544, hr2616, muhammad.shafique\}@nyu.edu
}

\begin{document}

\maketitle

%%%%%%%%%%%%%%%%%%%%%%%%%%%%%%%%%%%%%%%%%%%%%%%%%%%%%%%%%%%%%%%%%%%%%%%%%%%%%
%%%%%%%%%%%%%%%%%%%%%%%%%%%%%%%%%%%%%%%%%%%%%%%%%%%%%%%%%%%%%%%%%%%%%%%%%%%%%
\begin{abstract}
Continual learning (CL) in neural networks is a promising learning approach for updating the knowledge of deployed systems based on new data encountered at run time, that have not been learned during the offline training phase. 
To enable their power/energy-efficient CL processing, neuromorphic continual learning (NCL) concept with sparse event-based operations have recently been introduced. 
However, state-of-the-art NCL methods typically employ the memory-hungry data/experience replay, 
coupled with costly global learning rule (i.e., backpropagation-based surrogate gradient). 
To address these limitations, we propose \textit{\textbf{ENCL}}, \textit{a novel methodology that enables efficient NCL by employing a lightweight local learning rule and sub-network freezing strategy}. 
ENCL leverages the temporally- and spatially-local learning rule to perform efficient credit assignment for training new tasks, alleviating the expensive cost of global learning rule. 
ENCL also employs a sub-network freezing strategy based on the given tasks to avoid the needs for network architecture expansion, hence further improving the efficiency and scalability of the NCL training process.
To efficiently support this, we train each task with a small number of epochs as well as adjusted learning rate and timestep.
Experiment results on task-incremental learning (TIL) scenario show that, our ENCL consistently achieves strong performance across different datasets as compared to state-of-the-art works.
Specifically, ENCL achieves competitive accuracy comparable to that of the state-of-the-art data/experience replay, while reducing memory cost by up to 88\%, NCL training time by up to 51\%, average power consumption by up to 40\%, and energy consumption by up to 65\% across different benchmarking datasets (i.e., Split MNIST, Split Fashion MNIST, Split CIFAR-10, and Split CIFAR-100).
These results show that our ENCL is a promising approach toward energy-efficient CL for autonomous systems. 
\end{abstract}

% Uncomment the following to link to your code, datasets, an extended version or similar.
% You must keep this block between (not within) the abstract and the main body of the paper.
% Make sure that you do not de-anonymize yourself with these links.
% \begin{links}
%     \link{Code}{https://aaai.org/example/code}
%     \link{Datasets}{https://aaai.org/example/datasets}
%     \link{Extended version}{https://aaai.org/example/extended-version}
% \end{links}

%%%%%%%%%%%%%%%%%%%%%%%%%%%%%%%%%%%%%%%%%%%%%%%%%%%%%%%%%%%%%%%%%%%%%%%%%%%%%
%%%%%%%%%%%%%%%%%%%%%%%%%%%%%%%%%%%%%%%%%%%%%%%%%%%%%%%%%%%%%%%%%%%%%%%%%%%%%
\section{Introduction}
\label{Sec_Intro}

Neural network (NN)-based systems typically rely on the offline-based training to achieve high performance (e.g., accuracy) in the inference phase~\cite{Ref_LeCun_DeepLearning_Nature15}.
However, the knowledge of these systems may become outdated over time as they encounter new data (which usually represent new tasks) at run time, that have not been considered during their offline training phase. 
Consequently, this condition adversely affects the systems' performance (e.g., accuracy drop). 
Nowadays, it becomes important/critical for emerging application use-cases that demand adaptive capability in learning new task using new data from environments under low computational and memory budgets, such as resource-constrained autonomous systems (e.g., \add{UAVs, UGVs, and mobile robots})~\cite{Ref_Shaheen_CLsurvey_JIRS22}. 
However, performing a retraining process is not a practical solution as it requires a full dataset including data from both the old and new tasks, which is not always available/accessible due to the data restriction policy. 
Moreover, it also demands significant training time, computational and memory budgets, as well as power/energy consumption. 
Furthermore, retraining may suffer from \textit{catastrophic forgetting (CF)} issues, where the models quickly forget older tasks as learning a new task, if its process is not carefully performed~\cite{Ref_Minhas_Replay4NCL_DAC25}. 

\begin{figure}[t]
\centering
\includegraphics[width=\linewidth]{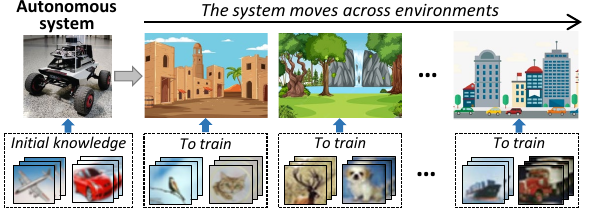}
\caption{An overview of the continual learning (CL) concept for a resource-constrained autonomous system.}
\label{Fig_CL}
\vspace{-0.3cm}
\end{figure}

To address these challenges, continual learning (CL) concept has emerged as a promising approach in enabling NN-based systems to adapt to new tasks, whose data samples are gathered directly from the operational environments~\cite{Ref_Zenke_CLintelligence_ICML17, RRef_DeLange_CLSurvey_TPAMI21, Ref_Wang_SurveyCL_TPAMI24}, as shown in Figure~\ref{Fig_CL}.
Many methods have been proposed in the literature for enabling CL capabilities in conventional artificial neural network (ANN) domain~\cite{Ref_Zenke_CLintelligence_ICML17, RRef_DeLange_CLSurvey_TPAMI21, Ref_Wang_SurveyCL_TPAMI24}. 
However, they typically incur high computational and memory costs due to their intensive operations, which stem from: (i) resource-hungry techniques (e.g., additional memory for data replay and expanding network architecture), and (ii) costly backpropagation-based learning process~\cite{Ref_Wang_SurveyCL_TPAMI24, Ref_Minhas_SurveyNCL_Access25}.
These limitations expose the needs for compute- and memory-efficient alternative designs to conventional CL methods. 
\textit{\textbf{The targeted problem} is, how can we develop an efficient CL method that incurs minimum computational and memory overheads}, thereby making it suitable and practical for on-device continual learning on low-power/energy-constrained systems. 

%%%%%%%%%%%%%%%%%%%%%%%%%%%%%%%%%%%%%%%
\smallskip
\textbf{Limitations of State-of-the-Art for Efficient CL:}
The targeted problem has led to increasing efforts in leveraging sparse event-based operations from spiking neural networks (SNNs) for devising highly efficient CL methods, i.e., so-called \textit{neuromorphic continual learning (NCL)}~\cite{Ref_Minhas_SurveyNCL_Access25}. 
The state-of-the-art NCL methods can be categorized based on their learning settings as follows.
\begin{itemize}
    \item \textit{Unsupervised learning}: 
    The existing works leverage the bio-plausible local learning rule, e.g., spike-timing dependent plasticity (STDP), to address CF through weight decay, adaptive threshold potential, adaptive learning rate, and adaptive synaptic plasticity~\cite{Ref_Panda_ASP_JETCAS18, Ref_Allred_CFN_FNINS20, Ref_Putra_SpikeDyn_DAC21, Ref_Putra_lpSpikeCon_IJCNN22}, and predictive coding~\cite{Ref_Antonov_CL4SNNs_NN22}.
    Despite their capabilities in handling unlabeled data, their benefits are limited to simple MNIST-level classification.
    \item \textit{Supervised learning}: 
    The existing works employ latent replay~\cite{Ref_Dequino_LatentReplay_ISVLSI24, Ref_Minhas_Replay4NCL_DAC25}, active dendrites~\cite{Ref_Pes_ActiveDendrites4SNNs_AICAS24}, network expansion~\cite{Ref_Han_DSDSNN_IJCAI23, Ref_Han_SORSNN_arXiv23}, parameter regularization~\cite{Ref_Antonov_CL4SNNs_NN22}, Bayesian~\cite{Ref_Skatchkovsky_BayesianCL_FNCOM22} and Hebbian~\cite{Ref_Xiao_HebbianCL_ICLR24} learning. 
    These works typically achieve higher performance (e.g., accuracy) than the unsupervised learning approach, but their global learning rule and training mechanism demand higher computational and memory overheads, and hence higher power/energy consumption. 
\end{itemize} 
This discussion highlights that, the state-of-the-art methods are not fully suitable for resource-constrained autonomous systems (e.g., mobile agents like UAVs/UGVs and robots). 
\textit{Therefore, there is a significant need for an alternative NCL method that offers effective performance with minimum computational and memory overheads}.

\begin{figure}[t]
\centering
\includegraphics[width=\linewidth]{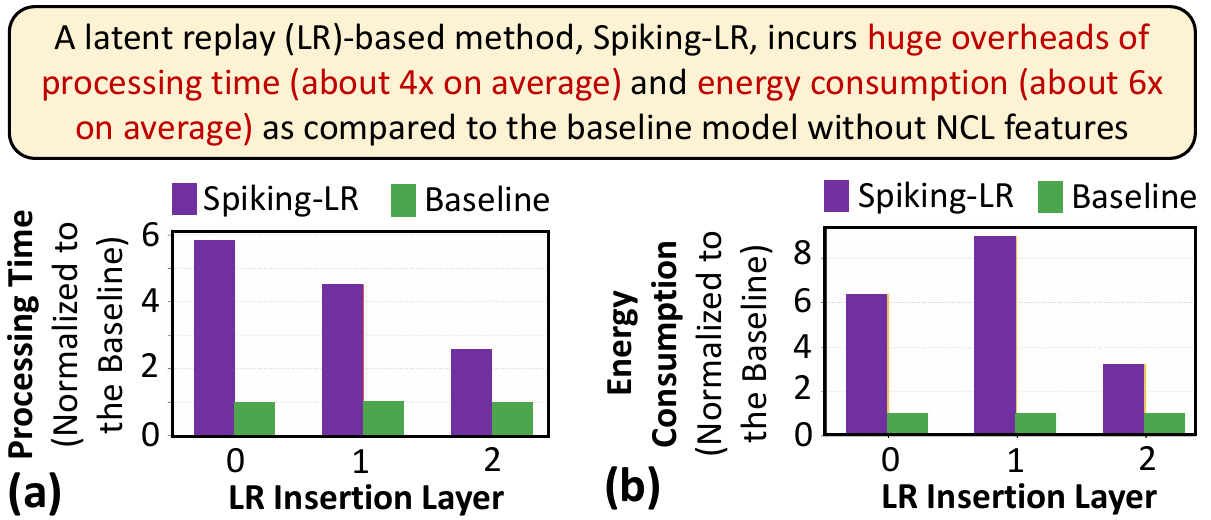}
\vspace{-0.3cm}
\caption{A latent replay (LR)-based method, e.g., Spiking-LR~\cite{Ref_Dequino_LatentReplay_ISVLSI24}, incurs significant \textbf{(a)} processing time/latency and \textbf{(b)} energy consumption, compared to the baseline model without NCL~\cite{Ref_Minhas_Replay4NCL_DAC25}.}
\label{Fig_Observe}
\vspace{-0.2cm}
\end{figure}

%%%%%%%%%%%%%%%%%%%%%%%%%%%%%%%%%%%%%%%
\smallskip
To illustrate the limitations of state-of-the-art NCL, a latent replay (LR)-based method Spiking-LR~\cite{Ref_Dequino_LatentReplay_ISVLSI24} is compared to the baseline without NCL techniques in terms of processing time (latency) and energy consumption. 
The comparison results are presented in Figure~\ref{Fig_Observe}, from which we observe that state-of-the-art NCL methods still incur notable additional compute and memory costs, leading to overheads in processing time and energy consumption.
We also identify the following associated research challenges for enabling efficient NCL in resource-constrained systems.
\begin{itemize}
    \item The solution should employ effective NCL algorithm so that high performance (e.g., accuracy) can be achieved over time across subsequent continual tasks. 
    \item The solution should efficiently perform the NCL training phase by minimizing the additional computational and memory overheads, hence optimizing the processing time and energy consumption.
\end{itemize}

\begin{figure*}[t]
\centering
\includegraphics[width=\linewidth]{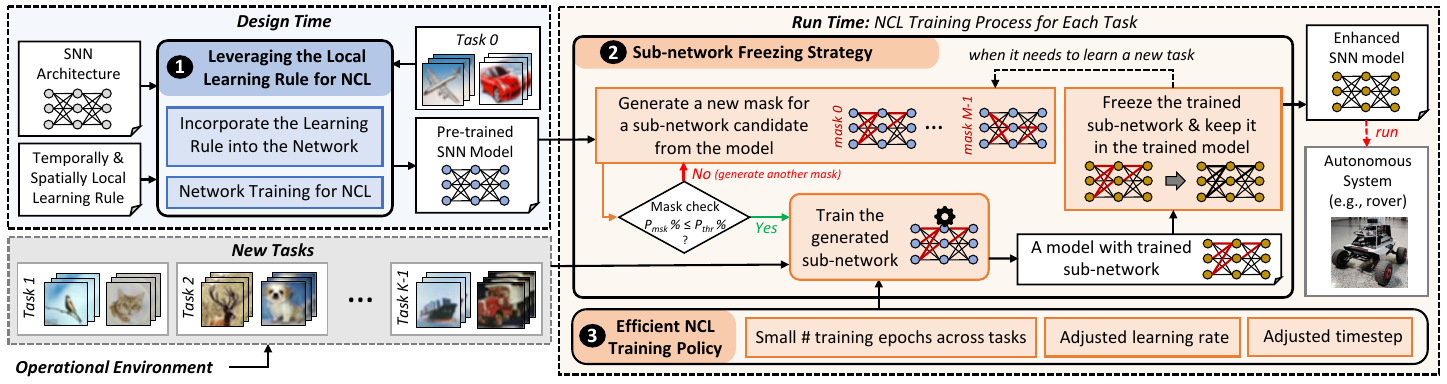}
\vspace{-0.5cm}
\caption{\add{Our ENCL methodology, showing its highlighted novel key steps: (1) leveraging the fully local learning rule, (2) devising a sub-network freezing strategy, and (3) designing an efficient NCL training policy.}}
\label{Fig_ENCL}
\vspace{-0.2cm}
\end{figure*}

%%%%%%%%%%%%%%%%%%%%%%%%%%%%%%%%%%%%%%%
\smallskip
\textbf{Our Novel Contributions:}
To address the targeted problem and challenges, we propose \textit{\textbf{ENCL}, a novel methodology for enabling \underline{E}fficient \underline{N}euromorphic \underline{C}ontinual \underline{L}earning by leveraging a local learning rule and sub-network freezing strategy}.
To achieve this, our ENCL employs the following key novel contributions (see the methodology in Figure~\ref{Fig_ENCL}).
\begin{itemize}
    \item \textbf{Leveraging the local learning rule for NCL:}
    We employ a temporally- and spatially-local learning rule to perform efficient credit assignment for training new tasks, hence substantially alleviating the expensive cost of global learning rule (e.g., backpropagation-based surrogate gradient learning) during the NCL training process. 
    \item \textbf{Sub-network freezing strategy:}
     \add{We improve efficiency and scalability of the NCL training process by assigning a task-specific sub-network to each task and protecting previously allocated parameters during subsequent training. Here, we generate a mask for the targeted task, train the corresponding sub-network, and then keep it in the saved dense model while preserving the task-specific mask. 
     } 
    \item \textbf{Efficient NCL training policy:}
    \add{We minimize the computational and memory costs during the NCL training process by training each new task with a relatively small number of epochs with adjusted learning rate and timestep.} 
\end{itemize}

Our ENCL methodology is evaluated using Python implementation that runs on a single Nvidia RTX 4090 Ti GPU, while considering the task-incremental learning (TIL) scenario.
\add{
Experimental results show that, ENCL achieves strong performance across different datasets as compared to state-of-the-art data replay.
Specifically, ENCL achieves comparable accuracy to the state-of-the-art data replay, while reducing memory cost by up to 88\%, NCL training time by up to 51\%, average power consumption by up to 40\%, and energy consumption by up to 65\% across different benchmarking datasets (i.e., Split MNIST, Split Fashion MNIST, Split CIFAR-10, and Split CIFAR-100)}
Therefore, our ENCL is a promising approach for realizing power/energy-efficient CL for autonomous systems.

%%%%%%%%%%%%%%%%%%%%%%%%%%%%%%%%%%%%%%%%%%%%%%%%%%%%%%%%%%%%%%%%%%%%%%%%%%%%%
%%%%%%%%%%%%%%%%%%%%%%%%%%%%%%%%%%%%%%%%%%%%%%%%%%%%%%%%%%%%%%%%%%%%%%%%%%%%%
\section{Related Work}
\label{Sec_Prelim}

%%%%%
%%%%%

%%%%%%%%%%%%%%%%%%%%%%%%%%%%%%%%%%%%%
\textbf{Neuromorphic Computing:}
Neuromorphic computing with SNN model leverages sparse event-based data representation and operations~\cite{Ref_Maass_SNN_NeuNet97, Ref_Putra_FSpiNN_TCAD20}. 
Its recent advancements in hardware~\cite{Ref_Akopyan_TrueNorth_TCAD15, Ref_Davies_Loihi_MM18, Ref_Neckar_Braindrop_IEEE19} and software~\cite{Ref_Rathi_PruneQuantizeSNN_TCAD18, Ref_Putra_SpikeNAS_TAI25} have shown ultra-low power/energy processing for diverse applications~\cite{Ref_Putra_SNNonCNP_IJCNN25}. 
A learning rule in SNN domain defines how information is learned by the model, and the existing rules in can be categorized as follows.
\begin{itemize}
    \item \textit{Global learning rule}:
    It performs model training under supervised settings. 
    The prominent approach is the surrogate gradient learning~\cite{Ref_Neftci_Surrogate4SNN_IEEESPM19}, such as spike layer error reassignment in time (SLAYER) \cite{Ref_Shresta_SLAYER_NeurIPS18}, spatio-temporal backpropagation (STBP)~\cite{Ref_Wu_STBP_FNINS18}, 
    and backpropagation through time (BPTT)~\cite{Ref_Fang_BPTT4SNN_ICCV21, Ref_Apolinario_STLLR_TMLR25}.
    Despite delivering high performance, this approach incurs huge computational and memory overheads due to costly global optimization during backpropagation. 
    \item \textit{Local learning rule}: 
    It usually leverages phenomena observed in neuroscience (e.g., STDP), that work locally in each synapse under unsupervised settings. 
    They usually offer high efficiency but slightly lower performance than supervised-based rules.
    Another work proposes an STDP-inspired temporally local learning rule~\cite{Ref_Apolinario_STLLR_TMLR25}. 
    A recent work proposes a temporally- and spatially-local learning rule (TESS) under supervised settings~\cite{Ref_Apolinario_TESS_IJCNN25}, which offers comparable performance and higher efficiency than the global learning rules.
    \textit{Hence, in this work, we select TESS as the learning rule for enabling efficient NCL}. 
\end{itemize}

%%%%%%%%%%%%%%%%%%%%%%%%%%%%%%%%%%%%%
\textbf{Local Learning Rule for SNNs:}
Recently, the temporally- and spatially-local learning rule for SNNs (TESS)~\cite{Ref_Apolinario_TESS_IJCNN25} has been proposed to address both spatial and temporal credit assignments by using locally available signals in each spiking neuron, as the following.
\begin{itemize}
    \item \textit{Temporal credit assignment using eligibility traces}:
    Eligibility traces are variables that encode the temporal history of synapses based on the pre- and post-synaptic activities. 
    These traces are used to identify if the weights of respective synapses should be updated, hence addressing the temporal credit assignment challenge.   
    \item \textit{Spatial credit assignment using local learning signals}: 
    A learning signal is needed for modulating the eligibility traces and addressing the spatial credit assignment challenge. 
    This signal is generated from the difference between the projected vector of output spikes in each layer and the target vector.   
    \item Weight updates are computed based on the learning signals ($m[t]$) and the eligibility traces ($e[t]$), and can be expressed as Eq.~\ref{Eq_TESS}. 
    Here, $w$ denotes the weight, $l$ denotes the layer, and $t$ denotes the timestep. 
    \begin{equation}
        \Delta w^l[t] = m[t] \cdot e[t]
    \label{Eq_TESS}
    \end{equation}
\end{itemize}

\begin{figure}[h]
% \vspace{-0.2cm}
\centering
\includegraphics[width=\linewidth]{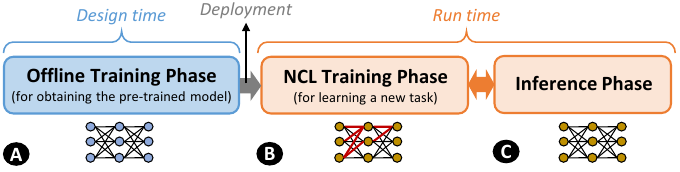}
\vspace{-0.3cm}
\caption{The NCL pipeline includes: the offline training phase at design time, the NCL training phase at run time, and the inference phase at run time.}
\label{Fig_NCLpipeline}
% \vspace{-0.3cm}
\end{figure}

%%%%%%%%%%%%%%%%%%%%%%%%%%%%%%%%%%%%%
\textbf{\add{NCL Pipeline and Scenarios:}}
\add{There are three phases considered in the NCL pipeline: \circledB{A} the offline training phase at design time, \circledB{B} the NCL training phase at run time, and \circledB{C} the inference phase at run time; see Figure~\ref{Fig_NCLpipeline}.}
NCL scenarios can be categorized based on the task identity (TID)~\cite{Ref_Minhas_SurveyNCL_Access25}. 
In this work, we consider the task incremental learning (TIL) scenario since it has been widely-used in the SNN community as the standard NCL evaluation.
In TIL, task boundaries are clearly defined, TID is known during the training and inference phases, and each task typically consists of multiple classes.

%%%%%%%%%%%%%%%%%%%%%%%%%%%%%%%%%%%%%%%%%%%%%%%%%%%%%%%%%%%%%%%%%%%%%%%%%%%%%
%%%%%%%%%%%%%%%%%%%%%%%%%%%%%%%%%%%%%%%%%%%%%%%%%%%%%%%%%%%%%%%%%%%%%%%%%%%%%
\section{The ENCL Methodology}
\label{Sec_ENCL}

Our ENCL methodology enables efficient NCL by \circledB{1} leveraging a fully local learning rule, \circledB{2} developing a sub-network freezing strategy, and \circledB{3} devising an efficient NCL training policy.
Its key steps are shown in Figure~\ref{Fig_ENCL}, and discussed further in the following sub-sections. 

%%%%
\begin{algorithm}[t]
\caption{Our offline training strategy under TIL}
\label{Alg_OfflineTrain}
\begin{algorithmic}[1]
\footnotesize
\renewcommand{\algorithmicrequire}{\textbf{INPUT:}}
\renewcommand{\algorithmicensure}{\textbf{OUTPUT:}}
\REQUIRE \textbf{(1)} The given SNN architecture ($net_{in}$); 
\textbf{(2)} Allocated connections-per-task ($c$); 
\textbf{(3)} Number of the targeted tasks ($K$); 
\textbf{(4)} Dataset of the targeted tasks ($data$); 
\textbf{(5)} Learning rate for the offline training ($\eta$): value ($\eta.val$) and scheduler ($\eta.{sched}$); 
\textbf{(6)} Training epochs for the offline training ($\epsilon$); \\
\ENSURE The trained model ($net_{out}$); \\
\smallskip
\textbf{BEGIN} \\
\smallskip
\textbf{Process}: \\
  \FOR{($k$=0; $k$$<$$K$; $k$++)}
    \IF{($k$==0)}
      \STATE $net_{tmp}$ = $net_{in}$; \\
    \ENDIF \\
    \STATE $mask$[$k$] = generate($net_{tmp}$, $c$); 
    \STATE $subnet$[$k$] = get\_subnet($net_{tmp}$, $mask$[$k$]); 
    \STATE $subnet$[$k$] = train($subnet$[$k$], $data$[$k$], $\eta$, $\epsilon$); 
    \STATE $net_{tmp}$ = freeze($net_{tmp}$, $subnet$[$k$]); 
  \ENDFOR \\
  \STATE $net_{out}$ = $net_{tmp}$;
  \RETURN $net_{out}$; \\
\textbf{END}
\end{algorithmic}
\end{algorithm}
\setlength{\textfloatsep}{6pt}
%%%%

%%%%%%%%%%%%%%%%%%%%%%%%%%%%%%%%%%%%%%%
\subsection{Leveraging the Fully Local Learning Rule for NCL}
\label{Sec_ENCL_LocalRule}

At design time, we integrate the temporally- and spatially-local learning rule (TESS)~\cite{Ref_Apolinario_TESS_IJCNN25} into the given network. 
We train this network to learn the known tasks during the offline training phase, while allocating space for accommodating the NCL training phase; see \circledB{A} in Figure~\ref{Fig_NCLpipeline}. 
To achieve this, we propose a training strategy for accommodating TIL.
Here, we allocate a specific sub-network for handling each given task, then train it with the training samples from the respective task.  
Its key steps are summarized in Algorithm~\ref{Alg_OfflineTrain} and discussed below.
\begin{enumerate}
    \item[\textbf{a1}] 
    \add{We generate a mask that occupies $1/K$ of the network to indicate a specific sub-network for handling the given task (with $K$ denotes the number of maximum accommodated tasks), thus leaving spaces for handling new tasks.} 
    Then, we go to \textbf{step-a2}.
    \item[\textbf{a2}] We train the sub-network with the training samples from the given task.
    \add{Since TIL provides the task identity and each sub-network is assigned to a specific task, we compute the local learning signals within the current task's label space instead of the full label space, so that the local objective is aligned with the task-specific pathway being trained.}
    Then, we go to \textbf{step-a3}.
    \item[\textbf{a3}] 
    \add{We record the parameters allocated to the current task, and keep the trained sub-network and its task mask in the model. 
    During later-task training, the recorded parameters from previous tasks are frozen and protected by gradient masking. 
    If we need to train the model for multiple tasks, then we go to \textbf{step-a1} for training another task; otherwise, we go to \textbf{step-a4}.}
    \item[\textbf{a4}] We conclude the process by employing the saved model as the fully trained model for deployment. 
\end{enumerate}
In this manner, a specific task (which potentially has multiple classes) is handled by a specialized sub-network. 
Furthermore, available spaces are also provided for accommodating new task learning at run time without network expansion.

%%%%%%%%%%%%%%%%%%%%%%%%%%%%%%%%%%%%%%%
\subsection{The NCL Training Mechanism}
\label{Sec_ENCL_NCL}

The deployed model often requires updating its knowledge based on new tasks encountered at run time (i.e., during the NCL training phase); see \circledB{B} in Figure~\ref{Fig_NCLpipeline}.
Each new task comes in the form of new data obtained from the operational environments.
On the other hand, the underlying hardware platforms usually have tight computational and memory budgets, thus making it difficult to perform effective NCL training. 
To address this, \textit{we propose an efficient NCL training mechanism that employs a sub-network freezing strategy with an improved training policy to effectively learn a new task}. 

%%%%%%%%%%%%%%
\smallskip

\textbf{Sub-Network Freezing Strategy:}
\add{We aim to generate a task-specific mask for each task and train the corresponding sub-network under a fixed hyperparameter setting. 
The trained sub-network is later evaluated considering its task mask. 
Specifically, we employ the following key steps when encountering a new task.
Its key steps are summarized in Algorithm~\ref{Alg_Freezing} and discussed below.}
\begin{enumerate}
    \item[\textbf{b1}]
    \add{We allocate a task-specific sub-network for each task by randomly generating a specific mask. 
    This allocated slot occupies a fixed fraction $1/K$ of the total capacity.  
    The mask for each task should be unique, but it may contain overlapping parts with masks from other tasks, hence it may not be sufficient for effectively learning a new task. 
    To address this, we employ \textit{a masking threshold} ($P_{thr}$\%) to decide if the generated mask should be used or replaced by a new mask, as discussed below.
    \begin{itemize}
        \item If the generated mask has a maximum $P_{msk}$\% overlap with one or more of previous masks, and it is higher than our defined threshold ($P_{msk}$\% > $P_{thr}$\%), then we replace some of the overlapping weights with other non-overlapping ones uniformly for each layer, so that this mask has a maximum $P_{thr}$\% overlap.
        \item If $P_{msk}$\% $\leq$ $P_{thr}$\%, then we continue with the generated mask directly.
    \end{itemize}
    Then, we go to \textbf{step-b2}.}
    \item[\textbf{b2}] 
    \add{We fully train the investigated sub-network with the training samples from the targeted task.
    Here, there are two possible conditions.}
    \begin{itemize}
        \item \add{\textit{If it is the first investigated sub-network}, we train and update all weights in the sub-network, then \textit{freeze} the fully trained weights and keep them in the model.} 
        \item \add{\textit{If it is not the first investigated sub-network}, we only train and update the non-overlapping weights in the sub-network, since the current mask may have overlaps with the frozen weights from previous masks. 
        Then, we \textit{freeze} the updated weights and keep them in the model.}
    \end{itemize}
    \add{Afterward, we go to \textbf{step-b1} to evaluate another mask, if applicable; otherwise, we go to \textbf{step-b3}.}
    \item[\textbf{b3}]
    \add{We update the record of parameters and store the task-specific mask for evaluation. 
    These saved masks allow each task to be evaluated with its corresponding sub-network.
    Then, we conclude the process by employing the saved model as the enhanced trained model for inference.}
\end{enumerate}

%%%%%%%%%%%%%%%
\smallskip

\textbf{Efficient NCL Training Policy:}
To minimize the computational and memory overheads during the NCL training phase while offering high performance, we devise an efficient training policy tailored for the proposed sub-network freezing strategy, as described in the following.
\begin{itemize}
    \item \textit{Small number of NCL training epochs} ($\hat{\epsilon}$):
    We employ a relatively small number of NCL training epochs $\hat{\epsilon}$ for learning each task, to reduce the compute and memory costs as well as balance the training process across tasks.
    \item \add{\textit{Adjusted learning rate} ($\hat{\eta}$):} 
    We employ an adjusted learning rate $\hat{\eta}$ with a fixed scheduling across the training epochs.
    These adjustments allow us to perform more aggressive weight updates, hence enabling fast adaptation to a new task. 
    In this manner, the limited performance associated with a small $\hat{\epsilon}$ can be alleviated.
    \item \textit{Adjusted timestep ($T$):}
    We adjust the timestep $T$ based on the dataset complexity, so that the network employs a longer timestep $T$ for a more complex dataset. 
\end{itemize}

 %%%%%%%%%%%%%%%
  \begin{algorithm}[t]
  \caption{\add{Our sub-network freezing strategy and training policy for each task under TIL}}
  \label{Alg_Freezing}
  \footnotesize
  \begin{algorithmic}[1]
  \renewcommand{\algorithmicrequire}{\textbf{INPUT:}}
  \renewcommand{\algorithmicensure}{\textbf{OUTPUT:}}
  \REQUIRE \add{\textbf{(1)} The current dense SNN model ($net$); 
  \textbf{(2)} The used-parameter record from previous tasks ($used$); 
  \textbf{(3)} Data samples of the new task-$k$ ($data[k]$); 
  \textbf{(4)} Per-task capacity fraction ($d$); 
  \textbf{(5)} Learning rate for the NCL training phase ($\hat{\eta}$); 
  \textbf{(6)} Training epochs for the NCL training phase ($\hat{\epsilon}$).} \\
  \ENSURE \add{The updated dense SNN model ($net_{out}$), task-$k$ mask ($mask[k]$), and updated used-parameter record ($used_{out}$).} \\
  \smallskip
  \textbf{BEGIN} \\
  \smallskip
  \textbf{Process}: \\
    \STATE \add{$mask[k]$ = generate\_mask($net$, $d$);} \\
      \STATE \add{$avail$ = $\neg\, used$;} \\
      
      \STATE \add{$mask[k]$ = $mask[k] \cap avail$;} \\
      
      \STATE \add{$subnet[k]$ = get\_subnet($net$, $mask[k]$);} \\
    \FOR{\add{($e$=0; $e$$<$$\hat{\epsilon}$; $e$++)}}
      \FOR{\add{each mini-batch $(x,y)$ in $data_k$}}
        \STATE \add{$loss$ = train($subnet_k$, $x$, $y$, $\hat{\eta}$);} \\
        \STATE \add{$subnet[k].grad$ = mask\_grad($subnet[k].grad$, $used$);} \\
        \STATE \add{$subnet[k]$ = update($subnet[k]$, $\hat{\eta}$);} \\
        \STATE \add{$subnet[k]$ = apply\_mask($subnet[k]$, $mask[k]$);} \\
      \ENDFOR
    \ENDFOR
    
    \STATE \add{store\_mask($mask[k]$);} \\
    
    \STATE \add{$used_{out}$ = update\_used\_params($used$, $mask[k]$);} \\
    \STATE \add{$net_{out}$ = save\_config($net$, $subnet[k]$, $mask[k]$);} \\
    \RETURN \add{$net_{out}$, $mask[k]$, $used_{out}$;} \\
  \textbf{END}
  \end{algorithmic}
  \end{algorithm}
  \setlength{\textfloatsep}{10pt}
%%%%

%%%%%%%%%%%%%%%%%%%%%%%%%%%%%%%%%%%%%%%%%%%%%%%%%%%%%%%%%%%%%%%%%%%%%%%%%%%%%
%%%%%%%%%%%%%%%%%%%%%%%%%%%%%%%%%%%%%%%%%%%%%%%%%%%%%%%%%%%%%%%%%%%%%%%%%%%%%
\section{Experiments}
\label{Sec_Exp}

To evaluate our ENCL methodology, we use a Python implementation and execute it on a single Nvidia RTX 4090 Ti GPU device. 
\add{The evaluation considers different benchmarking datasets, including: 
Split MNIST, Split Fashion MNIST, Split CIFAR-10, and Split CIFAR-100.
Each 10-class dataset (Split MNIST, Split Fashion MNIST, and Split CIFAR-10) is split into 5 tasks with 2 classes-per-task; while Split CIFAR-100 is split into 10 tasks with 10 classes-per-task.
The evaluation outputs include accuracy, NCL training memory and time, and power/energy consumption of NCL training.
}

\add{\textbf{NCL Scenario:} 
Each benchmark is organized to facilitate NCL process under the TIL scenario.
To do this, each benchmarking dataset is first split into a sequence of tasks: task 0, task 1, ..., task ($K$-1); with $K$ the number of maximum accommodated tasks. 
In the offline training phase, task 0 is learned to represent the initial knowledge of the model when deployed for the first time.
In the NCL training phase, the remaining tasks are learned sequentially in order.
}

\textbf{Comparison Partners:}
For comparison partners, we use the BPTT learning rule without any NCL method (BPTT-noNCL), the TESS learning rule without any NCL method (TESS-noNCL), and the BPTT learning rule with state-of-the-art data/experience replay (BPTT+ER).
Since ER~\cite{Ref_Rolnick_ER_NeurIPS19} is originally a CL method from ANN domain, in this evaluation, we adapt and implement it in SNN domain considering the BPTT learning rule.

\textbf{Network Architecture and Settings:}
To provide fair comparisons, we implement different CL methods on the same network architecture from~\cite{Ref_Apolinario_TESS_IJCNN25}: the VGG9 network with 8 convolutional (Conv) and 1 fully-connected (FC) layers. 
To accommodate the TIL scenario, we structure the last layer so that each task is associated with a specific FC head. 
Our sub-network freezing strategy in ENCL employs $P_{thr}$=5\% and $T$=5 for the 5-task benchmarks (Split MNIST, Split Fashion MNIST, and Split CIFAR-10), as well as $P_{thr}$=1\% and $T$=9 for Split CIFAR-100.
This network is trained by all NCL methods with a learning rate $\eta$=0.005 and 8 epochs-per-task for Split MNIST, Split Fashion MNIST, and Split CIFAR-10; and a learning rate $\eta$=0.001 and 100 epochs-per-task for Split CIFAR-100.

%%%%%%%%%%%%%%%%%%%%%%%%%%%%%%%%%%%%%%%%%%%%%%%%%%%%%%%%%%%%%%%%%%%%%%%%%%%%%
%%%%%%%%%%%%%%%%%%%%%%%%%%%%%%%%%%%%%%%%%%%%%%%%%%%%%%%%%%%%%%%%%%%%%%%%%%%%%
\section{Results and Discussion}
\label{Sec_Results}

\begin{figure*}[t]
\centering
\includegraphics[width=\linewidth]{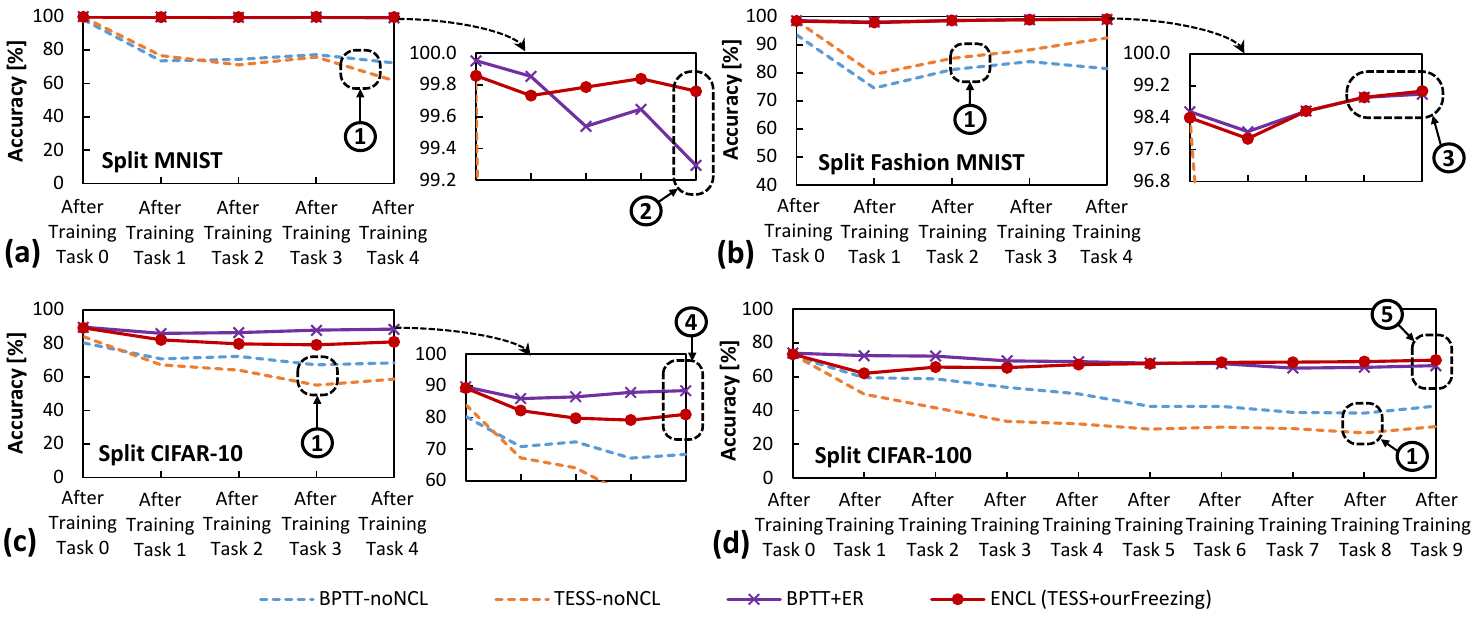}
\vspace{-0.5cm}
\caption{\add{Results for accuracy achieved by different learning methods (i.e., BPTT-noNCL, TESS-noNCL, state-of-the-art BPTT+ER, and our ENCL: TESS+ourFreezing) under the TIL scenario for \textbf{(a)} Split MNIST, \textbf{(b)} Split Fashion MNIST, \textbf{(c)} Split CIFAR-10, and \textbf{(d)} Split CIFAR-100.}}
\label{Fig_Results_Acc_SameArch}
\vspace{-0.2cm}
\end{figure*}

%%%%%%%%%%%%%%%%%%%%%%%%%%%%%%%%%%%%%%%
\subsection{Maintaining High Accuracy across Tasks}
\label{Sec_Results_Accuracy}

Experimental results for accuracy across tasks and datasets are presented in Fig.~\ref{Fig_Results_Acc_SameArch}. 
These results demonstrate that, BPTT-noNCL faces significant decreases in accuracy when learning new tasks.
In general, similar accuracy profiles are observed for TESS-noNCL, as shown by label \circled{1}.
These observations indicate that, BPTT-noNCL and TESS-noNCL suffer from catastrophic forgetting, leading to unstable behavior to maintain high accuracy.
The reason is that, BPTT-noNCL and TESS-noNCL do not have any NCL techniques or mechanisms to address the negative impact of subsequent task learning, hence making them prone to catastrophic forgetting and degrading accuracy.

To address this limitation, we may couple the BPTT learning rule with the state-of-the-art data/experience replay technique (i.e., BPTT+ER).
The experimental results show that, BPTT+ER can achieve good accuracy for different tasks across different benchmarking datasets as compared to the baselines (BPTT-noNCL and TESS-noNCL); see labels \circled{2}-\circled{5}.
The reasons can be attributed to the following. 
\begin{itemize}
    \item BPTT optimizes the global loss function to guide weight updates during the NCL training process, hence leading to effective weight tuning. 
    \item Data/experience replay mechanism effectively helps BPTT by presenting data from previous tasks to preserve old knowledge.
\end{itemize}
However, this high performance comes at high cost due to its intensive backpropagation-based training and data/experience replay mechanism, which will be discussed further in the following sub-sections (i.e., about memory cost, NCL training time, and power/energy consumption). 

These results also show that, our ENCL preserves strong accuracy profiles across tasks and datasets, highlighting its generality across different settings; see labels \circled{2}-\circled{5}.
ENCL achieves a slightly higher accuracy after training the last task than BPTT+ER by about 0.5\% for Split MNIST, by about 0.1\% for Split Fashion MNIST, and by about 3\% for Split CIFAR-100; see labels \circled{2}, \circled{3}, and \circled{5}.
For Split CIFAR-10, ENCL achieves a slightly lower accuracy after training the last task than BPTT+ER; see label \circled{4}.
The reasons can be attributed to the following.
\begin{itemize}
    \item ENCL employs a learning rule that updates weights based on local learning signals and local eligibility traces during the NCL training process, hence leading to effective and efficient weight tuning. 
    \item Our sub-network freezing strategy in ENCL effectively helps the local learning rule to handle subsequent task learning due to its task-specific pathway preservation within the network. 
\end{itemize}
Moreover, this competitive performance comes at low cost due to its lightweight local learning process and optimized NCL training policy, which will be discussed further in the following sub-sections (i.e., about memory cost, NCL training time, and power/energy consumption).

%%%
\begin{figure*}[t]
\centering
\includegraphics[width=\linewidth]{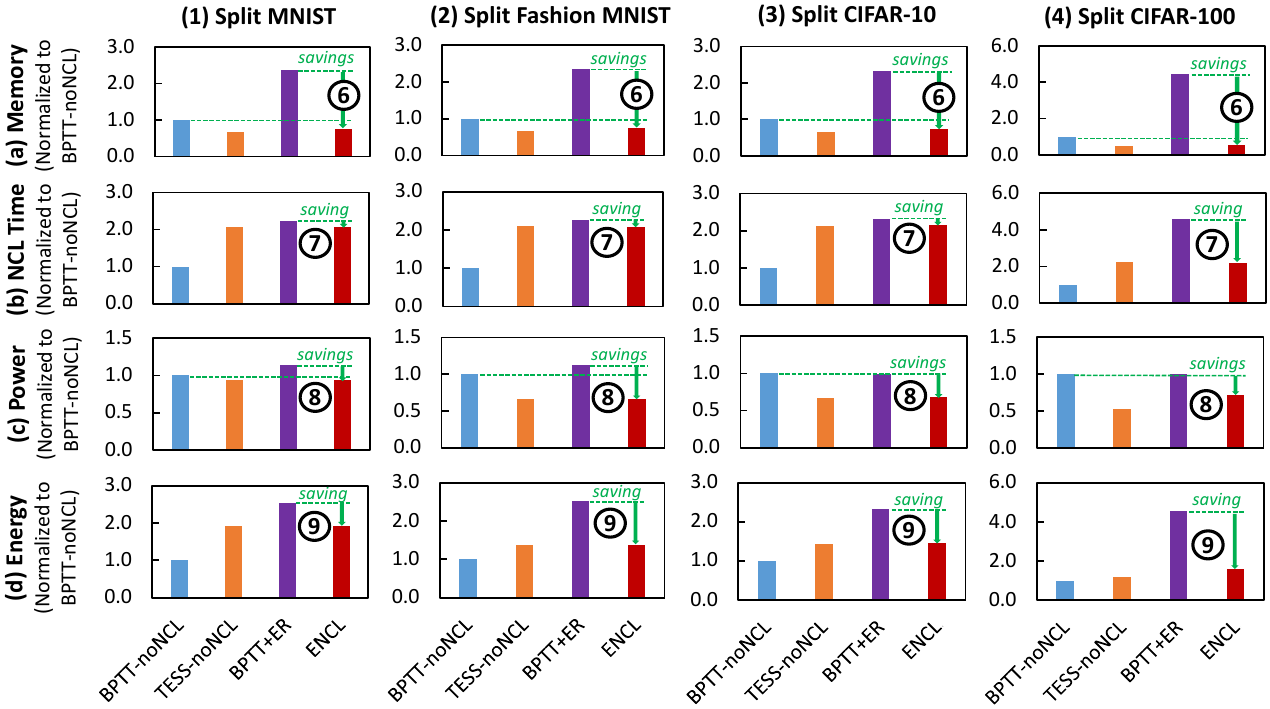}
\vspace{-0.5cm}
\caption{\add{Results for \textbf{(a)} memory cost, \textbf{(b)} NCL training time, \textbf{(c)} power consumption, and \textbf{(d)} energy consumption of NCL training, achieved by different learning methods (i.e., BPTT-noNCL, TESS-noNCL, state-of-the-art BPTT+ER, and ENCL) under the TIL scenario for different datasets (i.e., Split MNIST, Split Fashion MNIST, Split CIFAR-10, and Split CIFAR-100).}}
\label{Fig_Results_Resources_SameArch}
% \vspace{-0.2cm}
\end{figure*}
%%%

%%%%%%%%%%%%%%%%%%%%%%%%%%%%%%%%%%%%%%%
\subsection{Minimizing Memory Cost during NCL Training}
\label{Sec_Results_Memory}

Experimental results for the memory cost of NCL training process across tasks and datasets are presented in Fig.~\ref{Fig_Results_Resources_SameArch}(a). 
These results show that, TESS-noNCL incurs lower memory cost than BPTT-noNCL by about 32\%-53\% across datasets.
The reason is that, TESS-noNCL employs the learning rule that relies only on the available local signals, thereby incurring less memory cost than the backpropagation-based BPTT. 
Specifically, TESS has $Ln$ memory complexity, while BPTT has $TLn$; with $T$ is the total number of timesteps, $L$ is the number of layers in the network, and $n$ is the average number of neurons-per-layer~\cite{Ref_Apolinario_TESS_IJCNN25}.
Despite incurring relatively small memory costs, BPTT-noNCL and TESS-noNCL suffer from catastrophic forgetting. 
Adding the data/experience replay (ER) on top of the BPTT learning rule helps addressing the catastrophic forgetting issue, but it significantly increases the memory cost.
Specifically, BPTT+ER incurs 2.32$\times$-4.45$\times$ memory overheads as compared to BPTT-noNCL, highlighting the costly memory overhead required by the state-of-the-art solution.

These results also show that, our ENCL significantly reduces the memory cost of NCL training phase by 67\%-88\% as compared to the state-of-the-art BPTT+ER across different datasets; see label \circled{6}.
These memory savings can be associated with the following reasons. 
\begin{itemize}
    \item The memory complexity of the local learning rule in ENCL does not rely on the number of timesteps $T$~\cite{Ref_Apolinario_TESS_IJCNN25}, hence incurring smaller memory cost than the backpropagation-based BPTT.
    \item The sub-network freezing strategy in ENCL consumes a minor portion of memory due to its lightweight sub-network training process as compared to the costly ER.
\end{itemize}

%%%%%%%%%%%%%%%%%%%%%%%%%%%%%%%%%%%%%%%
\subsection{Minimizing NCL Training Time}
\label{Sec_Results_TrainTime}

Experimental results for the NCL training time across tasks and datasets are presented in Fig.~\ref{Fig_Results_Resources_SameArch}(b). 
These results show that, BPTT-noNCL has the fastest processing time as it does not employ any NCL techniques, thus having the lowest computational complexity but also suffering from catastrophic forgetting. 
TESS-noNCL incurs longer NCL training time than BPTT-noNCL since it has to compute learning signals and eligibility traces for each layer that are still not optimized for conventional underlying hardware platforms like GPUs.  
Meanwhile, BPTT+ER incurs 2.23$\times$-4.58$\times$ NCL training time overheads as compared to BPTT-noNCL, highlighting the costly time overhead required by the replay mechanism.

These results also show that, our ENCL reduces the NCL training time by 7\%-51\% as compared to the state-of-the-art BPTT+ER across different datasets; see label \circled{7}.
These NCL training time savings can be associated with the fact that, our sub-network freezing strategy in ENCL only considers a portion of the network when learning each task, which is smaller than the data/experience replay mechanism in BPTT+ER. 

%%%%%%%%%%%%%%%%%%%%%%%%%%%%%%%%%%%%%%%
\subsection{Optimizing Power and Energy Consumption during NCL Training Process}
\label{Sec_Results_PowerEnergy}

Experimental results for the power consumption of NCL training across tasks and datasets are presented in Fig.~\ref{Fig_Results_Resources_SameArch}(c). 
These results show that, TESS-noNCL incurs lower power than BPTT-noNCL by about 6\%-48\% across datasets.
The reason is that, TESS-noNCL employs TESS rule that relies only on the available local signals, which leads to smaller memory cost and hence less power consumption than BPTT, which relies on global optimization. 
Consequently, adding data/experience replay on top of the BPTT learning rule evidently increases power consumption.
Specifically, BPTT+ER incurs up to 1.14$\times$ the power consumption of BPTT-noNCL, highlighting the power overhead required by the state-of-the-art solution. 

These results also indicate that, ENCL effectively reduces the power consumption of NCL training phase by 17\%-40\% as compared to the state-of-the-art BPTT+ER.
These power savings can be associated with the following reasons.
\begin{itemize}
    \item The local learning rule in ENCL has smaller memory complexity than BPTT rule in BPTT+ER.  
    \item The NCL training policy in ENCL only considers a sub-network when learning each task, which is smaller than BPTT+ER, which always considers the entire network.
\end{itemize}
These power savings and reduced NCL training time (from previous sub-section) offered by our ENCL lead to significant energy savings as compared to the state-of-the-art BPTT+ER.
Specifically, ENCL effectively decreases the energy consumption of NCL training phase by 23\%-65\%.

%%%%%%%%%%%%%%%%%%%%%%%%%%%%%%%%%%%%%%%%%%%%%%%%%%%%%%%%%%%%%%%%%%%%%%%%%%%%%
%%%%%%%%%%%%%%%%%%%%%%%%%%%%%%%%%%%%%%%%%%%%%%%%%%%%%%%%%%%%%%%%%%%%%%%%%%%%%
\normalcolor
\section{Conclusion}
\label{Sec_Conclusion}

In this paper, we propose a novel ENCL methodology for enabling efficient NCL by leveraging a fully local learning rule to alleviate the expensive cost of global learning rule, and employing a sub-network freezing strategy to avoid network expansion. 
In TIL scenario, ENCL achieves strong performance across different datasets as compared to state-of-the-art data/experience replay: \add{obtaining comparable accuracy, while reducing memory cost by up to 88\%, NCL training time by up to 51\%, average power consumption by up to 40\%, and energy consumption by up to 65\% across different benchmarking datasets, including Split MNIST, Split Fashion MNIST, Split CIFAR-10, and Split CIFAR-100.}
These results highlight that our ENCL is a promising approach toward efficient and scalable CL for resource-constrained autonomous systems. 

%%%%%%%%%%%%%%%%%%%%%%%%%%%%%%%%%%%%%%%%%%%%%%%%%%%%%%%%%%%%%%%%%%%%%%%%%%%%%
%%%%%%%%%%%%%%%%%%%%%%%%%%%%%%%%%%%%%%%%%%%%%%%%%%%%%%%%%%%%%%%%%%%%%%%%%%%%%
\section*{Acknowledgments}
This work was partially supported by the NYUAD Center for Artificial Intelligence and Robotics (CAIR), funded by Tamkeen under the NYUAD Research Institute Award CG010. 

% \bigskip
% \noindent Thank you for reading these instructions carefully. We look forward to receiving your electronic files!

%%%%%%%%%%%%%%%%%%%%%%%%%%%%%%%%%%%%%%%%%%%%%%%%%%%%%%%%%%%%%%%%%%%%%%%%%%%%%
%%%%%%%%%%%%%%%%%%%%%%%%%%%%%%%%%%%%%%%%%%%%%%%%%%%%%%%%%%%%%%%%%%%%%%%%%%%%%
\bibliography{aaai2027} 

@inproceedings{Ref_Xiao_HebbianCL_ICLR24,
title={Hebbian Learning based Orthogonal Projection for Continual Learning of Spiking Neural Networks},
author={Mingqing Xiao and Qingyan Meng and Zongpeng Zhang and Di He and Zhouchen Lin},
booktitle={The Twelfth International Conference on Learning Representations (ICLR)},
year={2024},
}

@article{Ref_Shresta_SLAYER_NeurIPS18,
  title={Slayer: Spike layer error reassignment in time},
  author={Shrestha, Sumit B and Orchard, Garrick},
  journal={Advances in neural information processing systems},
  volume={31},
  year={2018}
}

@article{Ref_Neftci_Surrogate4SNN_IEEESPM19,
  title={Surrogate gradient learning in spiking neural networks: Bringing the power of gradient-based optimization to spiking neural networks},
  author={Neftci, Emre O and Mostafa, Hesham and Zenke, Friedemann},
  journal={IEEE Signal Processing Magazine},
  volume={36},
  number={6},
  pages={51--63},
  year={2019},
  publisher={IEEE}
}

@InProceedings{Ref_Fang_BPTT4SNN_ICCV21,
    author    = {Fang, Wei and Yu, Zhaofei and Chen, Yanqi and Masquelier, Timoth\'ee and Huang, Tiejun and Tian, Yonghong},
    title     = {Incorporating Learnable Membrane Time Constant To Enhance Learning of Spiking Neural Networks},
    booktitle = {Proceedings of the IEEE/CVF International Conference on Computer Vision (ICCV)},
    month     = {October},
    year      = {2021},
    pages     = {2661-2671}
}

@article{Ref_Wu_STBP_FNINS18,
  title={Spatio-temporal backpropagation for training high-performance spiking neural networks},
  author={Wu, Yujie and Deng, Lei and Li, Guoqi and Zhu, Jun and Shi, Luping},
  journal={Frontiers in neuroscience},
  volume={12},
  pages={331},
  year={2018},
  publisher={Frontiers Media SA}
}

@ARTICLE{Ref_Neckar_Braindrop_IEEE19,
  author={A. {Neckar} and S. {Fok} and B. V. {Benjamin} and T. C. {Stewart} and N. N. {Oza} and A. R. {Voelker} and C. {Eliasmith} and R. {Manohar} and K. {Boahen}},
  journal={Proceedings of the IEEE}, 
  title={Braindrop: A Mixed-Signal Neuromorphic Architecture With a Dynamical Systems-Based Programming Model}, 
  year={2019},
  volume={107},
  number={1},
  pages={144-164},}

@article{Ref_Apolinario_STLLR_TMLR25,
title={S-{TLLR}: {STDP}-inspired Temporal Local Learning Rule for Spiking Neural Networks},
author={Marco Paul E. Apolinario and Kaushik Roy},
journal={Transactions on Machine Learning Research},
issn={2835-8856},
year={2025},
note={}
}

@article{Ref_Skatchkovsky_BayesianCL_FNCOM22,
  title={Bayesian continual learning via spiking neural networks},
  author={Skatchkovsky, Nicolas and Jang, Hyeryung and Simeone, Osvaldo},
  journal={Frontiers in Computational Neuroscience},
  volume={16},
  pages={1037976},
  year={2022},
  publisher={Frontiers Media SA}
}

@article{Ref_Antonov_CL4SNNs_NN22,
  title={Continuous learning of spiking networks trained with local rules},
  author={Antonov, DI and Sviatov, KV and Sukhov, Sergey},
  journal={Neural Networks},
  volume={155},
  pages={512--522},
  year={2022},
  publisher={Elsevier}
}

@inproceedings{Ref_Dequino_LatentReplay_ISVLSI24,
  title={Compressed latent replays for lightweight continual learning on spiking neural networks},
  author={Dequino, Alberto and Carpegna, Alessio and Nadalini, Davide and Savino, Alessandro and Benini, Luca and Di Carlo, Stefano and Conti, Francesco},
  booktitle={2024 IEEE Computer Society Annual Symposium on VLSI (ISVLSI)},
  pages={240--245},
  year={2024},
  organization={IEEE}
}

@inproceedings{Ref_Pes_ActiveDendrites4SNNs_AICAS24,
  title={Active dendrites enable efficient continual learning in time-to-first-spike neural networks},
  author={Pes, Lorenzo and Luiken, Rick and Corradi, Federico and Frenkel, Charlotte},
  booktitle={2024 IEEE 6th International Conference on AI Circuits and Systems (AICAS)},
  pages={41--45},
  year={2024},
  organization={IEEE}
}

@article{Ref_Han_SORSNN_arXiv23,
  title={Adaptive reorganization of neural pathways for continual learning with spiking neural networks},
  author={Han, Bing and Zhao, Feifei and Pan, Wenxuan and Zhao, Zhaoya and Li, Xianqi and Kong, Qingqun and Zeng, Yi},
  journal={arXiv preprint arXiv:2309.09550},
  year={2023}
}

@inproceedings{Ref_Han_DSDSNN_IJCAI23,
author = {Han, Bing and Zhao, Feifei and Zeng, Yi and Pan, Wenxuan and Shen, Guobin},
title = {Enhancing efficient continual learning with dynamic structure development of spiking neural networks},
year = {2023},
isbn = {978-1-956792-03-4},
doi = {10.24963/ijcai.2023/334},
booktitle = {Proceedings of the Thirty-Second International Joint Conference on Artificial Intelligence},
articleno = {334},
numpages = {9},
location = {Macao, P.R.China},
series = {IJCAI '23}
}

@article{Ref_Shaheen_CLsurvey_JIRS22,
  title={Continual learning for real-world autonomous systems: Algorithms, challenges and frameworks},
  author={Shaheen, Khadija and Hanif, Muhammad Abdullah and Hasan, Osman and Shafique, Muhammad},
  journal={Journal of Intelligent \& Robotic Systems},
  volume={105},
  number={1},
  pages={9},
  year={2022},
  publisher={Springer}
}

@INPROCEEDINGS{Ref_Minhas_Replay4NCL_DAC25,
  author={Minhas, Mishal Fatima and Putra, Rachmad Vidya Wicaksana and Awwad, Falah and Hasan, Osman and Shafique, Muhammad},
  booktitle={2025 62nd ACM/IEEE Design Automation Conference (DAC)}, 
  title={Replay4NCL: An Efficient Memory Replay-based Methodology for Neuromorphic Continual Learning in Embedded AI Systems}, 
  year={2025},
  volume={},
  number={},
  pages={1-7},
  doi={10.1109/DAC63849.2025.11132839}}

@INPROCEEDINGS{Ref_Apolinario_TESS_IJCNN25,
  author={Apolinario, Marco P. E. and Roy, Kaushik and Frenkel, Charlotte},
  booktitle={2025 International Joint Conference on Neural Networks (IJCNN)}, 
  title={TESS: A Scalable Temporally and Spatially Local Learning Rule for Spiking Neural Networks}, 
  year={2025},
  volume={},
  number={},
  pages={1-9},
  doi={10.1109/IJCNN64981.2025.11227652}}

@article{RRef_DeLange_CLSurvey_TPAMI21,
  title={A continual learning survey: Defying forgetting in classification tasks},
  author={De Lange, Matthias and Aljundi, Rahaf and Masana, Marc and Parisot, Sarah and Jia, Xu and Leonardis, Ale{\v{s}} and Slabaugh, Gregory and Tuytelaars, Tinne},
  journal={IEEE transactions on pattern analysis and machine intelligence},
  volume={44},
  number={7},
  pages={3366--3385},
  year={2021},
  publisher={IEEE}
}

@article{Ref_Wang_SurveyCL_TPAMI24,
  title={A comprehensive survey of continual learning: Theory, method and application},
  author={Wang, Liyuan and Zhang, Xingxing and Su, Hang and Zhu, Jun},
  journal={IEEE Transactions on Pattern Analysis and Machine Intelligence (PAMI)},
  year={2024},
  publisher={IEEE}
}

@inproceedings{Ref_Zenke_CLintelligence_ICML17,
  title={Continual learning through synaptic intelligence},
  author={Zenke, Friedemann and Poole, Ben and Ganguli, Surya},
  booktitle={International conference on machine learning},
  pages={3987--3995},
  year={2017},
  organization={PMLR}
}

@INPROCEEDINGS{Ref_Putra_SNNonCNP_IJCNN25,
  author={Putra, Rachmad Vidya Wicaksana and Wickramasinghe, Pasindu and Shafique, Muhammad},
  booktitle={2025 International Joint Conference on Neural Networks (IJCNN)}, 
  title={Enabling Efficient Processing of Spiking Neural Networks with On-Chip Learning on Commodity Neuromorphic Processors for Edge AI Systems}, 
  year={2025},
  volume={},
  number={},
  pages={1-8},
  doi={10.1109/IJCNN64981.2025.11228298}}

@ARTICLE{Ref_Putra_SpikeNAS_TAI25,
  author={Putra, Rachmad Vidya Wicaksana and Shafique, Muhammad},
  journal={IEEE Transactions on Artificial Intelligence (TAI)}, 
  title={SpikeNAS: A Fast Memory-Aware Neural Architecture Search Framework for Spiking Neural Network-based Embedded AI Systems}, 
  year={2025},
  volume={},
  number={},
  pages={1-12},
  doi={10.1109/TAI.2025.3586238}}

@article{Ref_Minhas_SurveyNCL_Access25,
  title={Continual Learning with Neuromorphic Computing: Foundations, Methods, and Emerging Applications},
  author={Minhas, Mishal Fatima and Putra, Rachmad Vidya Wicaksana and Awwad, Falah and Hasan, Osman and Shafique, Muhammad},
  journal={IEEE Access},
  year={2025},
  publisher={IEEE}
}

@INPROCEEDINGS{Ref_Putra_SpikeDyn_DAC21,
  author={Putra, Rachmad Vidya Wicaksana and Shafique, Muhammad},
  booktitle={58th ACM/IEEE Design Automation Conference (DAC)}, 
  title={SpikeDyn: A Framework for Energy-Efficient Spiking Neural Networks with Continual and Unsupervised Learning Capabilities in Dynamic Environments}, 
  year={2021},
  volume={},
  number={},
  pages={1057-1062},
  doi={10.1109/DAC18074.2021.9586281}}

@inproceedings{Ref_Putra_lpSpikeCon_IJCNN22,
  title={lpSpikeCon: Enabling Low-Precision Spiking Neural Network Processing for Efficient Unsupervised Continual Learning on Autonomous Agents},
  author={Putra, Rachmad Vidya Wicaksana and Shafique, Muhammad},
  booktitle={2022 International Joint Conference on Neural Networks (IJCNN)},
  pages={1--8},
  year={2022}
}

@ARTICLE{Ref_Putra_FSpiNN_TCAD20,
  author={R. V. W. {Putra} and M. {Shafique}},
  journal={IEEE Transactions on Computer-Aided Design of Integrated Circuits and Systems (TCAD)},
  title={FSpiNN: An Optimization Framework for Memory-Efficient and Energy-Efficient Spiking Neural Networks}, 
  year={2020},
  volume={39},
  number={11},
  pages={3601-3613}}

@inproceedings{Ref_Rolnick_ER_NeurIPS19,
 author = {Rolnick, David and Ahuja, Arun and Schwarz, Jonathan and Lillicrap, Timothy and Wayne, Gregory},
 booktitle = {Advances in Neural Information Processing Systems},
 title = {Experience Replay for Continual Learning},
 volume = {32},
 year = {2019}
}

@ARTICLE{Ref_Allred_CFN_FNINS20,
AUTHOR={Allred, Jason M. and Roy, Kaushik},   
TITLE={Controlled Forgetting: Targeted Stimulation and Dopaminergic Plasticity Modulation for Unsupervised Lifelong Learning in Spiking Neural Networks},      
JOURNAL={Frontiers in Neuroscience},      
VOLUME={14},      
PAGES={7},     
YEAR={2020},      
DOI={10.3389/fnins.2020.00007},      
ISSN={1662-453X},   
}

@ARTICLE{Ref_Akopyan_TrueNorth_TCAD15,
author="F. {Akopyan} and J. {Sawada} and A. {Cassidy} and R. {Alvarez-Icaza} and J. {Arthur} and P. {Merolla} and N. {Imam} and Y. {Nakamura} and P. {Datta} and G. {Nam} and B. {Taba} and M. {Beakes} and B. {Brezzo} and J. B. {Kuang} and R. {Manohar} and W. P. {Risk} and B. {Jackson} and D. S. {Modha}",
journal="IEEE Transactions on Computer-Aided Design of Integrated Circuits and Systems (TCAD)",
volume="34",
number="10",
pages="1537-1557",
title="TrueNorth: Design and Tool Flow of a 65 mW 1 Million Neuron Programmable Neurosynaptic Chip",
month="Oct",
year="2015",
ISSN="",}

@ARTICLE{Ref_Davies_Loihi_MM18,
author="M. {Davies} and N. {Srinivasa} and T. {Lin} and G. {Chinya} and Y. {Cao} and S. H. {Choday} and G. {Dimou} and P. {Joshi} and N. {Imam} and S. {Jain} and Y. {Liao} and C. {Lin} and A. {Lines} and R. {Liu} and D. {Mathaikutty} and S. {McCoy} and A. {Paul} and J. {Tse} and G. {Venkataramanan} and Y. {Weng} and A. {Wild} and Y. {Yang} and H. {Wang}",
journal="IEEE Micro",
title="Loihi: A Neuromorphic Manycore Processor with On-Chip Learning",
year="2018",
volume="38",
number="1",
pages="82--99",
ISSN="",
month="Jan",}

@article{Ref_LeCun_DeepLearning_Nature15,
  title={Deep learning},
  author={LeCun, Yann and Bengio, Yoshua and Hinton, Geoffrey},
  journal={Nature},
  volume={521},
  number={7553},
  pages={436--444},
  year={2015},
  publisher={Nature Publishing Group UK London}
}

@ARTICLE{Ref_Rathi_PruneQuantizeSNN_TCAD18,
author="N. {Rathi} and P. {Panda} and K. {Roy}",
journal="IEEE Transactions on Computer-Aided Design of Integrated Circuits and Systems (TCAD)",
title="STDP-Based Pruning of Connections and Weight Quantization in Spiking Neural Networks for Energy-Efficient Recognition",
year="2019",
volume="38",
number="4",
pages="668--677",
ISSN="",
month="April",}

@article{Ref_Maass_SNN_NeuNet97,
title = "Networks of spiking neurons: The third generation of neural network models",
journal = "Neural Networks",
volume = "10",
number = "9",
pages = "1659--1671",
year = "1997",
issn = "0893-6080",
author = "W. Maass",}

@ARTICLE{Ref_Panda_ASP_JETCAS18,
author="P. {Panda} and J. M. {Allred} and S. {Ramanathan} and K. {Roy}",
journal="IEEE Journal on Emerging and Selected Topics in Circuits and Systems",
title="ASP: Learning to Forget With Adaptive Synaptic Plasticity in Spiking Neural Networks",
year="2018",
volume="8",
number="1",
pages="51-64",
month="March",
doi="10.1109/JETCAS.2017.2769684",
ISSN="",}

%%%%%%%%%%%%%%%%%%%%%%%%%%%%%%%%%%%%%%%%%%%%%%%%%%%%%%%%%%%%%%%%%%%%%%%%%%%%%
%%%%%%%%%%%%%%%%%%%%%%%%%%%%%%%%%%%%%%%%%%%%%%%%%%%%%%%%%%%%%%%%%%%%%%%%%%%%%
% Check whether the conference requires a reproducibility checklist to be included in the paper.
% If so, you can uncomment the following line and ajust the path to include it.
% \newpage
% \clearpage
% \input{ReproducibilityChecklist.tex}

\end{document}